\documentclass[fleqn,10pt]{wlscirep}
\usepackage[utf8]{inputenc}
\usepackage[T1]{fontenc}
\title{The Most Important Features in Generalized Additive Models Might Be Groups of Features}

\author[1,*]{Tomas Bosschieter}
\author[2]{Luis Fran\c{c}a}
\author[2]{Jessica Wolk}
\author[3]{Yiyuan Wu}
\author[3]{Bella Mehta, M.D.}
\author[4]{Joseph Dehoney}
\author[5]{Orsolya Kiss}
\author[5]{Fiona C. Baker}
\author[6]{Qingyu Zhao}
\author[2]{Rich Caruana}
\author[4]{Kilian M. Pohl}

\affil[1]{Stanford University, Institute for Computational and Mathematical Engineering, Stanford, CA 94305, USA}
\affil[2]{Microsoft Research, Redmond, WA 98052, USA}
\affil[3]{Hospital for Special Surgery, New York City, NY 10021, USA}
\affil[4]{Stanford University, Department of Psychiatry \& Behavioral Sciences, Stanford, CA 94305, USA}
\affil[5]{SRI International, Menlo Park, CA 94025, USA}
\affil[6]{Cornell University, Weill Cornell Medicine, New York City, NY 10065, USA}

\affil[*]{Corresponding author: tomasbos@stanford.edu}


\begin{abstract}
While analyzing the importance of features has become ubiquitous in interpretable machine learning, the joint signal from a group of related features is sometimes overlooked or inadvertently excluded. Neglecting the joint signal could bypass a critical insight: in many instances, the most significant predictors are not isolated features, but rather the combined effect of groups of features.
This can be especially problematic for datasets that contain natural groupings of features, including multimodal datasets.
This paper introduces a novel approach to determine the importance of a group of features for Generalized Additive Models (GAMs) that is efficient, requires no model retraining, allows defining groups posthoc, permits overlapping groups, and remains meaningful in high-dimensional settings. Moreover, this definition offers a parallel with explained variation in statistics. We showcase properties of our method on three synthetic experiments that illustrate the behavior of group importance across various data regimes.
We then demonstrate the importance of groups of features in identifying depressive symptoms from a multimodal neuroscience dataset, and  study the importance of social determinants of health after total hip arthroplasty.  
These two case studies reveal that analyzing group importance offers a more accurate, holistic view of the medical issues compared to a single-feature analysis.
\end{abstract}
\begin{document}

\flushbottom
\maketitle
%
%
\thispagestyle{empty}

\section{Introduction}
As machine learning models become ever-more ubiquitous in healthcare, there is a growing necessity for model interpretability. While it has proven difficult to reach a consensus on the meaning of interpretability \cite{doshi2017towards}, one particularly common instrument is determining the importance of individual features (aka feature importance or feature attribution), which in and of itself also does not have one single definition. Common definitions of feature importance include the average contribution of a feature to prediction in a separable model \cite{lou2013accurate, hastie2017generalized, nori2019interpretml}, average impurity decrease in tree-based models \cite{perrier2015feature, louppe2013understanding}, and the average increase in the model's error or loss after shuffling the values of a feature (i.e., permutation feature importance) \cite{breiman2001random, paschali2022bridging}. 

Providing an interpretation based on the importance of individual features is often challenging due to feature correlation and plausibility issues \cite{bilodeau2024impossibility}. For a group of correlated features, their joint signal could be crucial for prediction, while the individual features in the group might be attributed low feature importance \cite{darst2018using}. As a result, there could be signal in groups of correlated features that might be lost when analyzing individidual features by themselves. Even worse, these features could be filtered out during feature selection and thereby diminish the signal for inference \cite{louppe2013understanding}. This could be particularly troublesome on data containing natural groupings of features, such as modalities in multimodal data. Furthermore, interactions between features (in a group) are not captured through feature importance \cite{lundberg2020local}.
One way to address these issues is via group importance (aka grouped feature importance), i.e., by grouping features together and determining the importance of the group \cite{plagwitz2022supporting, au2022grouped}.

Group importance has previously been proposed using LASSO \cite{yuan2006model}, the Penalty Decomposition \cite{zhang2017computing}, Shapley values \cite{lundberg2017unified}, and permutation testing \cite{gregorutti2015grouped, paschali2022bridging}. Unfortunately, most of these methods come with computational challenges, mainly originating from the need to retrain the underlying machine learning model or compute expensive gradients. To address these challenges, we propose a robust, versatile, and computationally efficient method for computing group importance for Generalized Additive Models (GAMs), a very broad class of interpretable models that includes linear, logistic, and generalized linear models.

GAMs with interactions \cite{hastie2017generalized, lou2013accurate} are of the form \mbox{$g(\mathbb{E}(y)) = f_1(x_1) + \cdots + f_k(x_k) + \sum_{i \neq j} f_{i,j}(x_i, x_j)$}, 
where $g$ is a link function, $y$ the target, $x_i$ are features, $\beta_0$ is an intercept, and $f_i(x_i)$ and $f_{i,j}(x_i,x_j)$ are shape functions (aka component functions). In this setting, the (individual) importance $I_{x_j}$ of a feature $x_j$ is typically computed as the average absolute contribution of the corresponding shape function $f_j(\cdot)$ \cite{nori2019interpretml}, i.e.,
\begin{equation}
   I_{x_j} =  \frac{1}{|\mathcal{T}|} \sum_{t\in \mathcal{T}} |f_{j}(t_j)|,
\end{equation}
where $t \in \mathcal{T}$ represents samples $t$ in training data $\mathcal{T}$. For a group of features $G = \{x_{i_1}, \ldots, x_{i_k}\}$, we extend this definition and propose
\begin{equation}
    I_G \coloneqq  \frac{1}{|\mathcal{T}|} \sum_{t\in \mathcal{T}} |f_{i_1} ( t_{i_1}) + \cdots + f_{i_k}(t_{i_k})|,
\end{equation}
with mathematical justification given in Section~\ref{sec:methods_group_imp}. This definition of group importance has several properties we consider valuable, including:
\begin{itemize}
    \itemsep0em
    \item groups can be defined posthoc (and thus do not affect the model) and may be overlapping,
    \item computing group importance is computationally inexpensive due to the separability of the GAM,
    \item it remains meaningful when there are a lot of (correlated) features, unlike, e.g., group permutations,
    \item for a group consisting of one feature, its group importance equals the feature's individual importance,
    \item adding useless features, i.e. random noise, to a group does not change its group importance (up to small error $\epsilon$),
    \item one group could be all features in the dataset, yielding a relative scale on which to measure the importance of features and feature groups akin to percent of variance explained in statistics, i.e., percent of total importance `explained' by a group of features.
\end{itemize}

We implement our method with an Explainable Boosting Machine (EBM), a GAM-based model that achieves state-of-the-art accuracy while providing interpretability \cite{lou2013accurate, nori2019interpretml}. EBMs have recently gained popularity for their application in fields such as concrete compressive strength \cite{liu2023concrete}, brain tumors \cite{charlton4164349comparing}, Parkinson's disease \cite{sarica2022introducing}, and revealing healthcare biases that result in significant loss of life \cite{lengerich2022death}.
We evaluate the properties of our definition of group importance on three synthetic experiments and two real-world datasets that contain natural groupings of features. In the first case study, we identify the depressive symptom of negative valence \cite{woody2015integrating, paschali2022bridging} on a neuroscience dataset consisting of measurements from functional brain networks and groups (aka domains) such as sleep, traumatic experiences, biographic data, and social support. In the second case study, we focus on the influence of community-level social determinants of health in predicting mortality for total hip arthroplasty using a large-scale healthcare dataset ($>100$k samples). These determinants, encompassing environmental, economic, and social conditions, are paramount in understanding health outcomes.
Our main contributions are as follows:
\begin{itemize}
    \itemsep0em
    \item Proposing a definition for group importance for Generalized Additive Models that is computationally inexpensive.
    \item Showing that group importance offers enhanced interpretability in addition to individual feature importance, especially for correlated features, as demonstrated on two real-world datasets.
    \item Demonstrating group importance can be used as a low-cost, yet effective method for feature selection, and defining a parallel between group importance and explained variance.
\end{itemize}

\section{Methods}\label{sec:methods}
We first describe Explainable Boosting Machines (EBMs) as a specific implementation of a Generalized Additive Model (GAM), and how feature importance is typically computed in this setting. Then, we extend this approach to define group importance. Subsequently, we turn to feature selection and briefly discuss the parallel between group importance and explained variation in statistics.


\subsection{Explainable Boosting Machine}\label{sec:EBM}
Explainable Boosting Machines (EBMs) \cite{lou2013accurate} are Generalized Additive Models (GAMs) \cite{hastie2017generalized} with interaction terms, i.e.,
\begin{equation}
    g(\mathbb{E}[y]) = \beta_0 + f_1(x_1) + \cdots + f_n(x_n) + \sum_{i \neq j} f_{i,j}(x_i,x_j)
\end{equation}
The shape functions $f_i(\cdot)$ and $f_{i,j}(\cdot, \cdot)$ are trained using cyclic gradient boosting with shallow learners \cite{lou2013accurate}. During training, a {\it purification process} \cite{nori2019interpretml, lengerich2020purifying} ensures that the contribution of a feature $x_k$ for prediction is not moved between $f_k(x_k)$ and its interaction terms $f_{k, \ell}(x_k, x_\ell)$ for $\ell \in \{1, \ldots, n\} \backslash \{k\}$. That is, $f_k(x_k)$ encodes the sole and total contribution of $x_k$ to prediction (i.e., the main effect), while $f_{k, \ell}(x_k, x_\ell)$ encodes solely the interaction term adjusted for all main effects. Thus, the shape functions form a functional ANOVA decomposition \cite{lengerich2020purifying}. For enhanced interpretability, these functions are typically zero-centered and can be visualized to show the contribution of each feature or interaction term to prediction.

We measure individual, `global' feature importance as the average \emph{absolute} contribution of a feature to prediction over all samples in the training dataset. Notably, feature importance can sometimes refer to \emph{local} feature importance, which attributes importance to features for one sample prediction specifically \cite{nori2019interpretml}, but we only consider global feature importance in this paper. That is, for feature $x_j$ and samples $t \in \mathcal{T}$ taking on values $(t_1, \ldots, t_n)$, the feature importance $I_{x_j}$ \cite{nori2019interpretml} of $x_j$ is typically defined as
\begin{equation}\label{eq:1ft_def_ftimp}
   I_{x_j} =  \frac{1}{|\mathcal{T}|} \sum_{t\in \mathcal{T}} |f_{j}(t_j)|.
\end{equation}
Note, each contribution $|f_j(t_j)|$ can include weights defined by the user \cite{nori2019interpretml}.
This approach not only provides insights into the internal mechanisms of the model, but also ensures a symmetric attribution of importance, i.e., features signaling very high or low risk receive similar importance (a direct consequence of the zero-centering in the shape functions).

\subsubsection{Correlated features} \label{sec:EBM_cor_features}
Note, as a design choice, EBMs split the shared signal of highly correlated features evenly. This ensures that each feature receives a proportionate amount of signal and importance. This is a property worth highlighting, as there also exist alternate approaches. For example, other GAM implementations might explicitly aim to assign all importance to one of the features rather than splitting it evenly, such as in GRAND-SLAMIN' \cite{ibrahim2023grand}.

\subsection{Group importance}\label{sec:methods_group_imp}
Similar to individual feature importance, group importance has been well-studied and knows many definitions. Group importance is understood differently across various domains, each with its own set of definitions and accompanying challenges \cite{plagwitz2022supporting, au2022grouped}. This lack of consensus might lead to arbitrary selection of definitions rather than standardization. We aim to bridge this gap for GAMs specifically and propose a definition that naturally extends the notion of individual feature importance.


\subsection{Group importance for GAMs}\label{sec:methods_group_imp_GAM}
To posthoc evaluate the importance of a group of features after training the GAM, we average the absolute values of the sum of feature contributions in a group over all samples in the training dataset $\mathcal{T}$. That is, let $G = \{x_{i_1}, \ldots, x_{i_k}\}$ be a set of $k$ features whose group importance $I_G$ we wish to evaluate. Then, we define group importance as 
\begin{equation}\label{eq:group_imp}
    I_G \coloneqq  \frac{1}{|\mathcal{T}|} \sum_{t\in \mathcal{T}} |f_{i_1} ( t_{i_1}) + \cdots + f_{i_k}(t_{i_k})|.
\end{equation}
This is a canonical choice given that
\begin{enumerate}
    \itemsep0em
    \item it is  a direct extension of the $k=1$ case, with the property that a feature's individual importance equals the importance of a group consisting of only that feature,
    \item the {\it null} group (i.e. group consisting of no features) has 0 importance,
    \item group importance satisfies $0 \leq I_{\{x_{i_1}, \ldots, x_{i_k}\}} \leq I_{x_{i_1}} + \cdots + I_{x_{i_k}}$.
\end{enumerate}

Observe that (iii) is a particularly important and useful property, as it represents the triangle inequality for grouped feature importance. As such, (iii) naturally extends to groups: if $G_1, \ldots, G_\ell$ are groups of features and we let $G = \{G_1, \ldots, G_\ell\}$, then \\
\mbox{$0 \leq I_G \leq I_{G_1} + \cdots I_{G_\ell}$}. This is a direct result of the definition in Eq.~\eqref{eq:group_imp}. Note, another natural extension of group importance would have been $|f_{i_1}(t_{i_1})| + \cdots + |f_{i_k}(t_{i_k})|$ as the summand. However, this would result in group importance simply being the sum of feature importances, as
\begin{align*}
    \tilde{I}_{\{x_{i_1}, \ldots, x_{i_k}\}} &= \frac{1}{|\mathcal{T}|} \sum_{t\in \mathcal{T}} (|f_{i_1}(t_{i_1})| + \cdots + |f_{i_k}(t_{i_k})|) \label{eq:scuffed_defn_group_imp} \\
    &= \frac{1}{|\mathcal{T}|} \sum_{t\in \mathcal{T}} |f_{i_1}(t_{i_1})| + \cdots + \frac{1}{|\mathcal{T}|} \sum_{t\in \mathcal{T}} |f_{i_k}(t_{i_k})| \\ 
    &= I_{x_{i_1}} + \cdots + I_{x_{i_k}}.
\end{align*}

Similar to other definitions of group importance \cite{gregorutti2015grouped, lundberg2017unified, yuan2006model, paschali2022bridging, zhang2017computing}, our definition is permutation invariant. That is, for any permutation $\sigma \in \text{Sym}(\{1, 2, \ldots, k\})$, the symmetry group, it immediately follows from the definition that $I_{\sigma(G)} = I_{G}$. Besides permutation invariance, our notion of group importance is also invariant under feature scaling: scaling the values of a feature correspondingly stretches the shape function, thus leaving feature and group importance invariant. In the same vein, having a different number of features between groups does not artificially inflate the larger group's importance, since all shape functions are zero-centered and we adopt the definition in Eq.~\eqref{eq:group_imp} rather than the (flawed) alternative definition above. 

\subsection{Explained variation and feature importance}
We now draw a parallel between group importance and \textit{explained variation} in statistics, which represents to what extent (a set of variables in) a model accounts for the variation (often variance) of a target. In regression problems, explained variance is typically defined as the correlation coefficient $R^2$, a common metric in general\cite{o1982measures}. For classification problems, however, it depends on the setting, and explained variation is generally considered a term interchangeable with feature importance, or other measures of `goodness of fit' \cite{ballabio2013classification}.

In Principal Component Analysis (PCA) \cite{wold1987principal}, the scale for explained variance is the sum of the individual component variances $\Var(x_1) + \cdots + \Var(x_n)$. For independent variables (e.g., principal components) $x_1, \ldots, x_n$, the variance explained by $x_i$ is then defined as $\frac{\Var(x_i)}{\Var(x_1) + \cdots + \Var(x_n)}$ \cite{shen2008sparse}. More generally, if the $x_i$ are independent or at least weakly correlated, $\Var(x_1) + \cdots + \Var(x_n)$ can be approximated by the total variance $\Var(x_1 + \cdots + x_n)$, which is parallel to the total importance $I_{x_1, \ldots, x_n}$ in a GAM, i.e., the importance of the group containing all features.
As such, our method of comparing the importance of a subset of features to the total importance is similar in behavior to the explained variance in statistics and particularly relevant for datasets with weakly correlated (ideally uncorrelated) features. Interestingly, when the contribution of features to a model are not independent, feature importance and variance explained are not necessarily fully additive: if `total importance' in a GAM were defined as $I_{x_1} + \cdots + I_{x_n}$, the maximum is generally not achieved unless all feature contributions are independent and additive, and thus is not an appropriate scale. In fact, in pathological settings, there could exist a group $G$ of features such that $I_{G} > I_{x_1, \ldots, x_n}$, unless further assumptions about learning optimality can be made.

Once relative group importances have been computed, or a ranking has been obtained more generally, this can also be used for feature selection. That is, by picking the top $k$ groups of features that are assigned most importance by the model. This is particularly useful when feature selection has to be performed at a group-level.


\section{Synthetic Experiments}\label{sec:synthetic_experiments}

To demonstrate the inner workings of our proposed definition of group importance for GAMs, we generate 3 synthetic datasets. We examine the behavior exhibited when features contain signal that is (1) additive and perfectly correlated, (2) opposing and independent, and (3) opposing and highly correlated. One real-world healthcare example of opposing (and correlated) signal can be found in age-related diseases: advanced age indicates increased risk, while a sufficient amount of medication lowers risk, therefore containing opposite signal. At the same time, age and the amount of medication are likely correlated.

Generalizing from these three extreme, synthetic cases, we then study group importance as a function of correlation in both the additive and opposing signal case. We also review various edge cases in Supplementary Information~S1. Note, the EBMs achieve near-perfect prediction accuracy in each synthetic experiment due to the simplistic setup of the synthetic data. The synthetic datasets consists of samples $\{(x_i, z_i, y_i)\}_{i=1}^{n=10^6}$, where $x_i~\overset{\text{iid}}{\sim}~\mathcal{U}[0,10]$ is sampled from the uniform distribution; we define $z_i$ and $y_i$ in each subsection. We denote $x = (x_1, \ldots, x_n)^T$ and $[n] \coloneqq \{1,2,\ldots, n\}$. 

\subsection{Additive signals}\label{sec:synth_additive_signal}
To create synthetic data consisting of additive signals, let $z$ be a copy of $x$. Furthermore, let $\epsilon_i \overset{\text{iid}}{\sim} \mathcal{N}(0,1)$ be noise sampled from the normal distribution in order to construct label $y_i = 1$ if $x_i + \epsilon_i > 5$ and $0$ else. No matter whether $z_i = x_i$ or $z_i = -x_i \; \forall i \in [n]$, $z$ and $x$ are equally indicative of any target $y$.

\begin{figure}[H]
    \centering
    \includegraphics[width=0.65\linewidth]{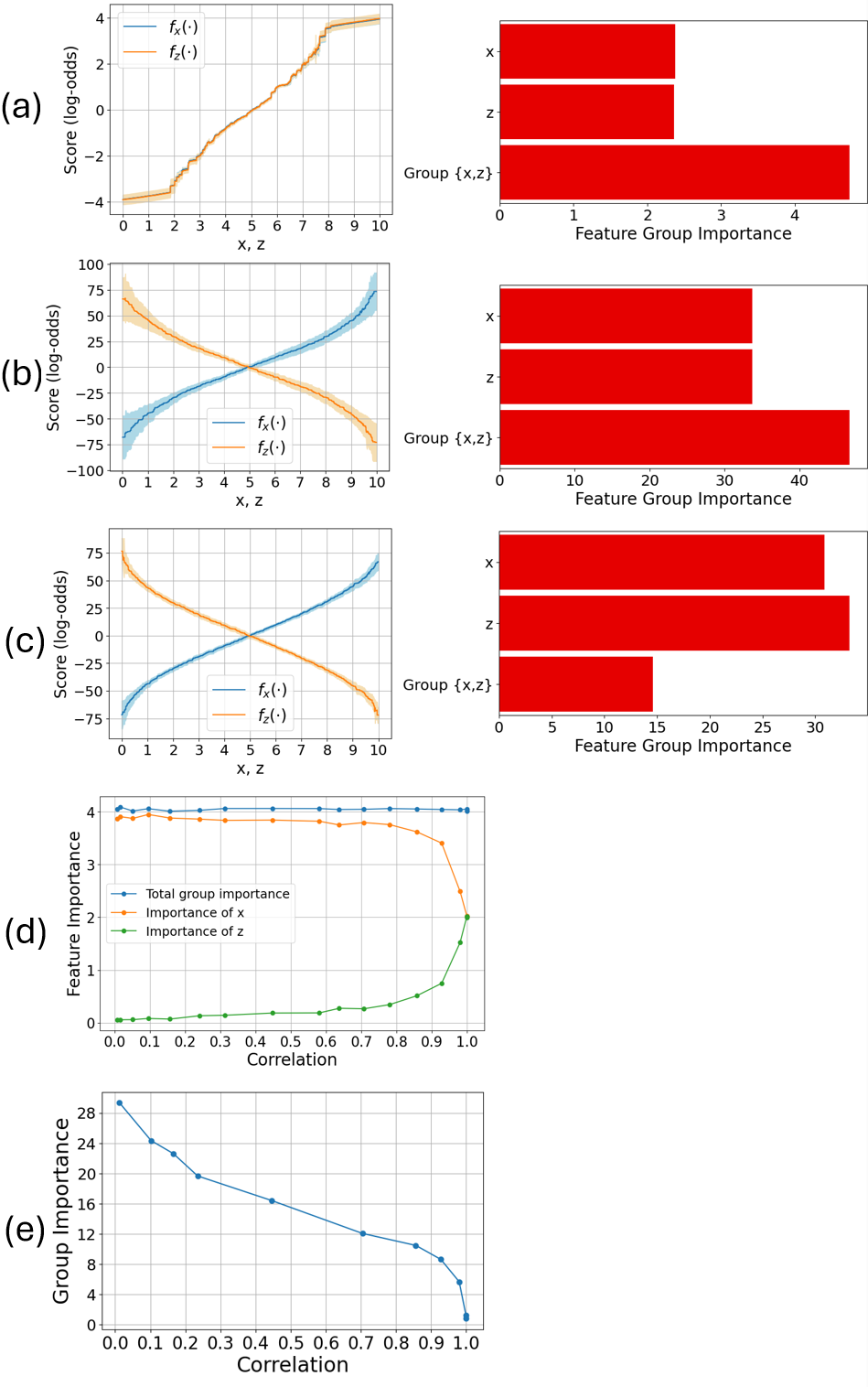}
    \caption{Group importance of the Explainable Boosting Machine (EBM) on various synthetic datasets with two variables x and z, and target y. By design, these variables can be either be additive or conflicting in predicting target y; that is, the shape functions of x and z could have similar or opposing shapes. Part (a) shows additive signal through z=x. Part (b) shows independent, conflicting signals. while part (c) shows correlated, conflicting signals. While the group importance ranking in (b) is larger than each single feature importance, the group importance in (c) shows that the two signals almost always cancel each other since the features x and z are highly correlated and close in value. Part (d) demonstrates group importance as a function of correlation for additive signals, while part (e) visualizes group importance as a function of correlation for conflicting signals. Note that the case of zero correlation in part (d) corresponds to the experiment in part (b). High correlation in part (e) corresponds to the experiment in part (c), and the extreme case of perfect corerlation in part (e) is explored in Supplementary Information Fig. S3.}
    \label{fig:synth_exp_all5}
\end{figure}

Then, due to perfect correlation, $f_x(\cdot)$ and $f_z(\cdot)$ each learn identical signal that is half of the total signal in the data (see Figure~\ref{fig:synth_exp_all5}(a)), as described in Section~\ref{sec:EBM_cor_features}. Thus, the group importance of $x$ and $z$ is equal to the sum of their individual importances, i.e., $I_{\{x, z\}} \approx 2I_{x}$ (see Figure~\ref{fig:synth_exp_all5}(a)).  Similarly, if a feature $x$ is duplicated and negated to $-x$, the corresponding shape functions $f_x(\cdot)$ and $f_{-x}(\cdot)$ contain negated, yet equivalent, signal and therefore also have equal feature importance. This is a direct consequence of the scaling property discussed previously. This example is shown in Supplementary Information~S1.1.

\subsection{Independent, conflicting signals}\label{sec:synth_independent_conflicting_signal}
To obtain a feature $z$ that is independent of $x$, sample $z_i~\overset{\text{iid}}{\sim}~\mathcal{U}[0,10]$. Then, to create opposing signal of $z$ with respect to $x$, let \\\mbox{$y_i = 1$} if $x_i > z_i$ and $0$ else. We visualize the shape functions $f_x(t_x)$ and $f_z(t_z)$ and show the individual and group importances in Figure~\ref{fig:synth_exp_all5}(b).

By symmetry in the construction of $y_i$, an increase in $x_i$ increases the probability that $y_i = 1$ (for random $z_i~\sim~\mathcal{U}[0,10]$), while an increase in $z_i$ \textit{decreases} that probability (for random $x_i~\sim~\mathcal{U}[0,10]$). As a result, the shape functions $f_x(\cdot)$ and $f_z(\cdot)$ in the EBM are each other's inverses, and their group importance is less than the sum of individual importances. 


To analyze the behavior shown in the group and feature importances in a more detailed fashion, recall that the group importance of features $x$ and $z$ is formulated as 
\begin{equation}\label{eq:conflicting_signal_exp2}
I_{\{x,z\}} = \frac{1}{|\mathcal{T}|} \sum_{t\in \mathcal{T}} |f_{x}(t_x) + f_{z}(t_{z})|.
\end{equation}
In this synthetic experiment, features $x$ and $z$ contain exactly opposite signal, i.e., $f_x(\cdot) \approx -f_z(\cdot)$. Because feature values $t = (t_x, t_z) \in \mathcal{T}$ are independently sampled, there might be many samples $t$ where $f_x(t_x) \gg 0$ and $f_z(t_z) \ll 0$ (and vice versa) such that these terms do not cancel each other out and, as a result, contribute to the group importance. Similarly, there are roughly equally many samples such that $f_x(t_x) \gg 0$ and $f_z(t_z) \gg 0$, as well as $f_x(t_x) \ll 0$ and $f_z(t_z) \ll 0$, assuming large $n$. Thus, the group importance $I_{x,z}$ is slightly larger than $I_x$ and $I_z$, but far smaller than their sum $I_x + I_z$.

\subsection{Correlated, conflicting signals} \label{sec:exp3_corr_conflicting_signals}
We now examine how our proposed definition of group importance behaves for correlated features that contain opposing signal. This is not an uncommon scenario in practice as, in many instances, features tend to be correlated and samples be drawn non-iid.

To have $z$ be correlated with $x$, we let $z_i = x_i + \delta_i$, where we sample $\delta_i \sim \mathcal{U}[-2, 2]$ (and scale $z_i$ to $[0,10]$), so that $x$ and $z$ are highly correlated. To retain conflicting signal, we again define $y_i = 1$ if $x_i > z_i$ and $0$ else.

The shape functions remain mostly unchanged with respect to the previous synthetic experiment, and the individual feature importances also remain balanced. However, in Figure~\ref{fig:synth_exp_all5}(c) we see that the group importance $I_{\{x,z\}} < I_x \approx I_z$. This is a direct result of the correlation and, in particular, $t_x \approx t_z$ for all samples $(t_x,t_z) \in \mathcal{T}$ and $f_x(t_x) \approx -f_z(t_z)$, meaning the shape functions nearly cancel each other out for most samples. That is, for perfectly opposing signal: the more correlation, the lower the group importance compared to the individual feature importances. In general, if feature $x_{i_1}$ is such that $I_{\{ x_{i_1}, x_{i_2}, \ldots, x_{i_k}\}} < I_{x_{i_1}}$, then the features $x_{i_2}, \ldots, x_{i_n}$ must contain opposing signal as a whole compared to $x_{i_1}$, since they lower the mean absolute contribution to the target.



\subsection{Group importance as function of correlation} \label{sec:generalized_data_setup}
In the three synthetic examples above, we considered the extreme cases of perfectly additive or conflicting signal with either high or no correlation between feature values. We now generalize these experimental setups by computing group importance as a function of the Pearson correlation between $x$ and $z$. This is achieved by varying boundary $b \in [0, 10^3]$ such that $\delta_i \sim \mathcal{U}[-b, b]$ yields a nearly uniformly distributed $z_i = x_i + \delta_i$ on one end, or a $z$ perfectly correlated with $x$ on the other end of the spectrum. (We again scale $z_i$ to $[0,10]$ for convenience in constructing $y$.) In other words, we vary the correlation between $x$ and $z$ and study the effects on their group importance, in both the additive and conflicting setup. 

\subsubsection{Additive signal}\label{sec:generalization_additive_signal}
To generalize the experiment on additive signals in Section~\ref{sec:synth_additive_signal}, we keep the same target $y$ and compute the group importance as a function of the Pearson correlation $\rho$ between $x$ and $z$ (see Figure~\ref{fig:synth_exp_all5}(d)). We also show the individual importances of $x$ and $z$ given the asymmetry in how $y$ is constructed.
Observe that the total group importance is constant (up to random noise) since target $y$ is determined using only feature $x$ and not $z$ (see Figure~\ref{fig:synth_exp_all5}(d)).
For perfect correlation (i.e., $\rho = 1$ as in Section~\ref{sec:synth_additive_signal}), we see that $x$ and $z$ are both assigned half the total, group importance, while $y$ is assigned near-zero importance when it is near-uniformly distributed and uncorrelated with $x$.

\subsubsection{Conflicting signal}
Generalizing the conflicting signals experiment of Section~\ref{sec:synth_independent_conflicting_signal}, we visualize the group importance as a function of the correlation in Figure~\ref{fig:synth_exp_all5}(e). 
We observe a roughly linear decline in the group importance of $x$ and $z$ as their correlation increases, transitioning to a more exponential decrease at higher correlation levels. This pattern likely arises from measuring group importance as the mean \emph{absolute} contribution, which inherently challenges the convergence to zero. However, the group importance does converge to zero as the correlation goes to one.

\section{Real-World Experiments}
We asses the computational efficiency of group importance and assess the utility and relevance of group importance in two real-world medical datasets. In our first case study, we identify the depressive symptom \emph{negative valence} based on groups of correlated features drawn from functional brain networks and various life and behavioral domains. As a second case study, we investigate the importance of social determinants of health in the outcome of hip replacements. For each we analyze the group importance ranking, and use this as a guide for applying group importance for low-cost feature selection purposes.

\subsection{Computational Efficiency}\label{sec:computational-efficiency}
\begin{figure}[H]
    \centering
    \includegraphics[width=0.75\linewidth]{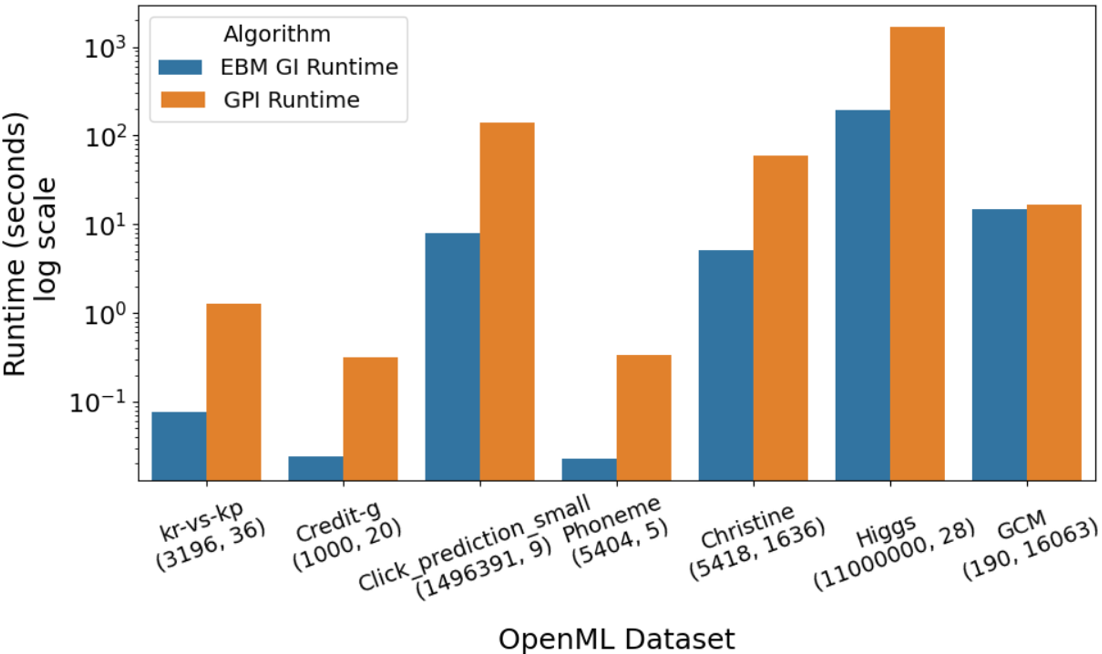}
    \caption{Computational cost in terms of runtime of various metrics of group importance. Our method is compared to Grouped Permutation Importance (GPI) \cite{plagwitz2022supporting} for a variety of datasets picked to represent different characteristics, such as number of data samples and variables. These seven datasets were downloaded from OpenML, an open-source platform for sharing datasets that is commonly used for benchmarking machine learning models\cite{OpenML2013}. We were unable to reproduce the runtimes for the permutations-based ranking \cite{paschali2022bridging}. Our method is an order of magnitude faster to compute on all OpenML datasets but GCM, where it is very marginally better. The experiments were performed on an Intel(R) Xeon(R) CPU @ 2.20GHz.}
    \label{fig:comp_cost_GroupImps}
\end{figure}

Unlike other group importance methods, our method is computationally low-cost. It requires only $k$ function evaluations for each sample $t \in \mathcal{T}$, resulting in a complexity of $\mathcal{O}(k|\mathcal{T}|)$, or $\mathcal{O}(|\textnormal{no. features in group}| \cdot |\textnormal{no. samples}|)$ in terms of function evaluations. This is an order of magnitude faster than the popular grouped permutation importance (GPI) method \cite{plagwitz2022supporting} across different experimental setups, even if the number of features or samples is very high. For various OpenML datasets with such different data regimes\cite{OpenML2013}, there is an order of magnitude difference in terms of runtime (on an Intel(R) Xeon(R) CPU @ 2.20GHz) of group importances implemented for GAMs versus GPI (see Figure~\ref{fig:comp_cost_GroupImps}).

\subsection{Neuroscience Experiments} \label{sec:neuro_experiments}
While better understanding the underlying mechanism of depressive symptoms could hugely improve mental health care, this is a difficult task as the manifestation of the symptoms is quite heterogeneous. That is, they are not confined to a single, easily identifiable brain region or network, nor clear changes in behavior \cite{iordan2017brain, lopresti2013review, binnewies2021associations}. 

\subsubsection{Data} \label{sec:real_world_data_MDD}
We conduct this experiment on publicly available data acquired by the National Consortium on Alcohol and Neurodevelopment in Adolescence (NCANDA) \cite{ncanda2021} (data release \verb|NCANDA_PUBLIC_6Y_REDCAP_V04| \cite{pohl2022ncanda_public_6y_redcap_v04} and, for fMRIs,\\ \verb|NCANDA_PUBLIC_6Y_RESTINGSTATE_V01| \cite{pohl2022ncanda_public_6y_restingstate_v01}). The NCANDA dataset contains multimodal and longitudinal data comprised of biographical, self-reported behavioral, and fMRI data of 1396 observations (aka visits) from 521 participants (ages 12 to 18 years).

\paragraph{Predictors}\label{sec:brain_networks}
There are 23 resting state fMRI (rs-fMRI) features as well as 131 demographic \& behavioral measurements that are categorized into eight disjoint groups \cite{paschali2022bridging, brown2015national}:
\begin{enumerate}
    \itemsep0em
    \item \textbf{Life \& Trauma Events} (full: \textit{Life events and childhood trauma}):  encompasses trauma, emotional neglect, and (parental) marital separation
    \item \textbf{Personality Traits}: include agreeableness, emotional stability, extraversion, positive thinking, and cognitive restructuring.
    \item \textbf{Neuropsychological Battery}: includes problem-solving and reasoning skills, attention, memory, and language \cite{zgaljardic2010neuropsychological}.
    \item \textbf{Executive Function Spectrum} (full: \textit{Behavior Rating Inventory of Executive Function} (BRIEF)): assesses executive functions inhibition and shifting, measuring the ability to resist or delay impulses and to tolerate change \cite{gioia2000test}.
    \item \textbf{Social Support Features}: records relationships with friends and family, as well as active participation in social clubs and events.
    \item \textbf{Sleep Patterns}: tracks wake-up time and bedtime for weekdays and weekends, as well as sleepiness and circadian preferences.
    \item \textbf{Demographics} (full: \textit{Demographics and pubertal development}): includes age, sex, BMI, ethnicity, pubertal development scale score.
    \item \textbf{Alcohol \& Drug Use Features}: records alcohol and substance use.
\end{enumerate}
The 23 brain functional measurements were generated by processing each resting-state fMRI (acquired on a GE or Siemens scanner; voxels:  64x64x32, voxel size: 3.75mm x 3.75mm x 5mm; TR: 2.2s) using the NCANDA pipeline outlined in \cite{zhao2019longitudinally}. The pipeline consists of skull-stripping, motion correction, detrending, spatial smoothing, and anomaly detection through a manual quality check. All scans of this analysis had a minimum of 7.5 minutes of usable scan time and were registered to the SRI24 atlas \cite{rohlfing2010sri24}. 23 intrinsic function networks (IFN) were extracted via  longitudinal independent component analysis (ICA), and each IFN is a brain network that regulates different aspects of our cognition and behavior, such as memory, attention, emotions, and motor skills. Each of the 23 IFNs is assigned a score representing its strength, i.e. the average functional connectivity within the brain network, sometimes referred to as the functional network's \emph{efficiency score} \cite{muller2018influences}. To account for scanner differences in the acquisition (GE or siements), we regress out scanner type from the efficiency scores by performing scanner-level standardization. 

\paragraph{Target variable} We identify \textit{negative valence} \cite{woody2015integrating, paschali2022bridging}, an aggregate statistic for feeling fear, anxiety, sadness, loss, and threat, as a measure of depressive symptoms. It has been shown that negative valence's subdomains are highly correlated with Major Depressive Disorder (MDD) and that MDD reflects dysfunctions in negative valence systems \cite{medeiros2020positive}. Negative valence has a prevalence of 6.02\% across all visits.

\subsubsection{Group Importance Ranking}\label{sec:group_imp_ranking}
To assess the accuracy achieved by the EBM, we conducted stratified 10-fold cross validation. All visits by a subject were consistently in either the training or test set in order to avoid information leakage to the test set. This is necessary since the data is longitudinal in nature. The EBM model attains a mean test AUC of $0.874 \pm 0.011$, balanced accuracy of $0.735 \pm 0.016$, and Brier score of $0.061 \pm 0.002$ across visits, where the errors show the standard error of the mean (SEM). This is on par with models such as random forests, recurrent neural networks, and logistic regression trained on this dataset \cite{paschali2022bridging}. The strong predictive performance of the EBM model ensures the robustness of its findings. Emotional stability, a feature belonging to the Personality Traits group, was assigned the highest individual feature importance.

\begin{figure}[H]
    \centering
    \includegraphics[width=0.75\linewidth]{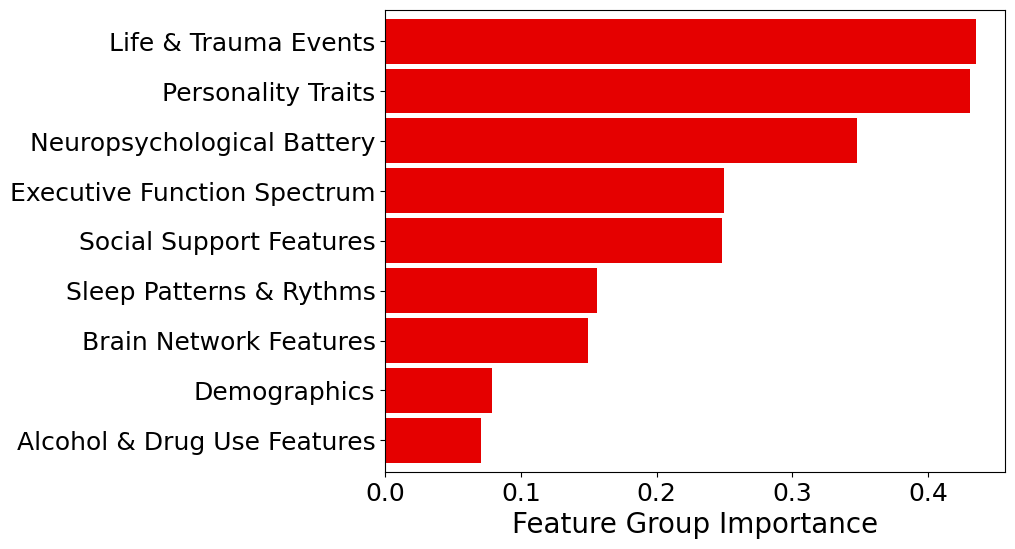}
    \caption{Group importance scores computed using our method for identifying depressive symptoms using data from the National Consortium on Alcohol and Neurodevelopment in Adolescence (NCANDA) \cite{ncanda2021, pohl2022ncanda_public_6y_redcap_v04}. The groups Life \& Trauma Events and Personality Traits are the most predictive groups, highlighting the significance of behavioral and environmental factors.}
    \label{fig:teaser}
\end{figure}

To gain a deeper insight into which \emph{groups of features} were driving prediction, we combined all fMRI features into one group and the rest of the measurements into the eight groups described in Section~\ref{sec:real_world_data_MDD}. This yields the group importances shown in Figure~\ref{fig:teaser}, revealing that five of the nine groups are more important than the most important individual feature. 

Interestingly, the most important group is Life \& Trauma Events while only five of its features are among the 20 most important features. The group Life \& Trauma Events contains mostly highly correlated features, namely chronic/non-chronic life aspects that are controllable/uncontrollable and have a positive/negative/ambiguous outcome (each of the $2\cdot 2 \cdot 3 = 12$ combinations is a feature). Naturally, these features are highly correlated and are thus each attributed relatively low feature importance. Yet, when considered collectively, these features form the most critical group for prediction, highlighting the significance of group importances in our analysis. This phenomenon of high group importance for correlated, low-importance features was also observed in the additive synthetic experiment (Section~\ref{sec:synth_additive_signal}) and its generalization (Section~\ref{sec:generalization_additive_signal}).


Table~\ref{tab:comparing_group_rankings} presents the rankings of the NCANDA groups based on their group importance, as determined by several methods. Our method's ranking is compared to those obtained through Grouped Permutation Importance (GPI) \cite{plagwitz2022supporting} and a previously reported ranking on the NCANDA dataset, derived using another permutation-based method \cite{paschali2022bridging}. While there is significant overlap in rankings, our method appears to attribute much more importance to neuropsychology battery than the permutations-based ranking \cite{paschali2022bridging}, ranking it as the third most important group instead of the seventh. GPI ranks neuropsychology as the fifth most important group. Our higher ranking seems more in line with the literature, which extensively links depressive symptoms to measurements of the neuropsychological battery group, including reasoning skills \cite{engle2005cognitive, hu2022depression}, memory \cite{burt1995depression}, and intellectual functioning \cite{brown1994cognitive, clark1985intellectual}. All three methods agree on the two most important groups, namely Life \& Trauma Events and Personality Traits, while Executive Function is also found to be an important group. On the other end of the spectrum, the group Demographics is attributed low group importance across the board, with its individual features also being attributed low individual feature importance. Similarly, both our method and GPI rank the group Alcohol \& Drug Use as the least important group, while it is the 5th most important group according to the permutations-based ranking \cite{paschali2022bridging}.

One group omitted by Paschali et al. (2022) \cite{paschali2022bridging} was Brain Networks. Given that functional interactivity between brain networks tend to be correlated \cite{marrelec2006partial}, just like in the case of Life \& Trauma Events, this decision might have been based on analyzing low feature importances. However, our analysis reveals that the importance of the group of fMRI features is substantial and is larger than all individual features but emotional stability, while also outranking the two groups Demographics and Alcohol \& Drug Use.

\subsubsection{Feature selection}
\begin{figure}[H]
    \centering
    \includegraphics[width=0.75\linewidth]{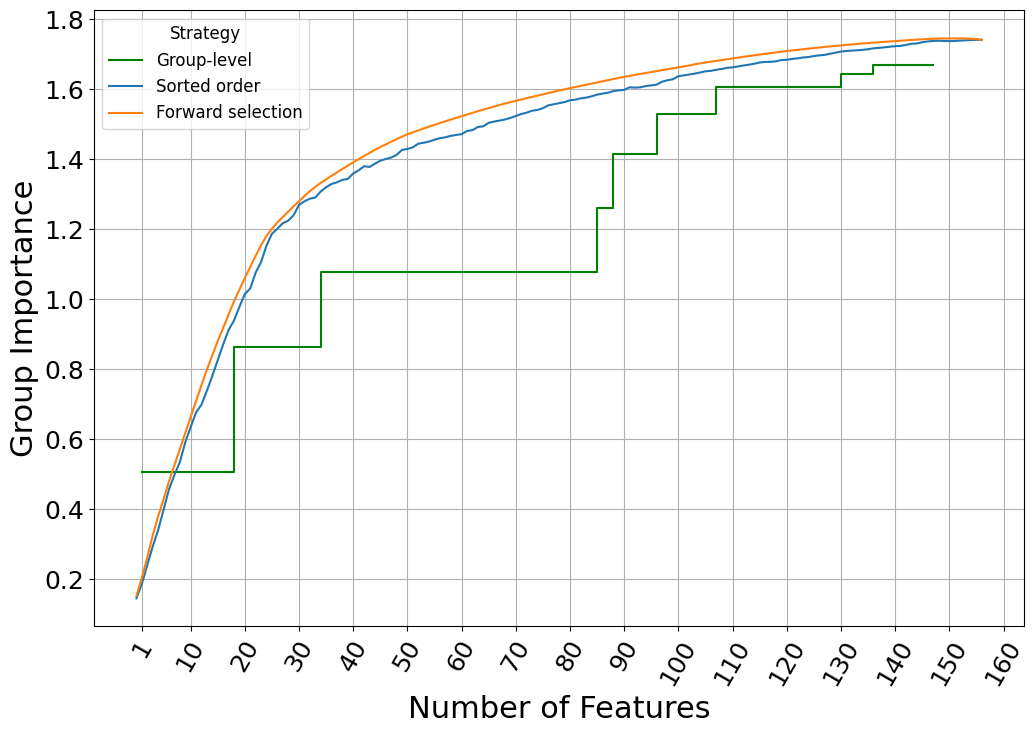}
    \caption{Cumulative group importance when successively adding more feature groups. We compare our group-level ranking to both the sorted order of individual feature importances and a greedy forward selection strategy, which is a standard method for feature selection. Our method tracks the greedy forward selection strategy quite closely while only adding one group at a time, suggesting feature selection through performing forward selection on groups of features may be almost as effective as performing feature selection on individual features, when maximizing for group importance.}
    \label{fig:cum_importance}
\end{figure}

Following the analysis of the group importance ranking, our next step is to evaluate the practicality of using this ranking for feature selection. We conduct a series of experiments where we first train the EBM using the top $1, \ldots, 9$ groups, assessing model accuracy with test AUC and the Brier score through 5-fold cross-validation. The cumulative importance of these groups as they are sequentially added in order of their importance can be found in Figure~\ref{fig:cum_importance}.

\begin{figure}[H]
    \centering
    \begin{subfigure}[t]{0.48\linewidth}
        \centering
        \includegraphics[width=\linewidth]{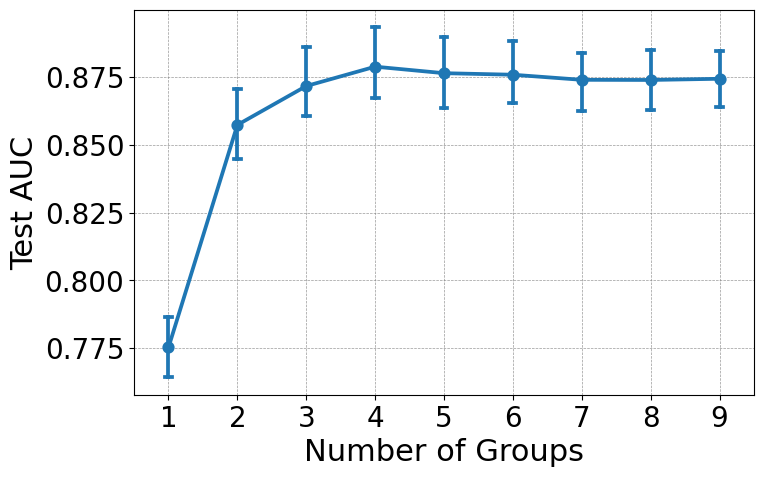}
        \caption{Test AUC (higher is better).}
        \label{fig:auc_x_numgroups}
    \end{subfigure}%
    ~ 
    \begin{subfigure}[t]{0.48\linewidth}
        \centering
        \includegraphics[width=\linewidth]{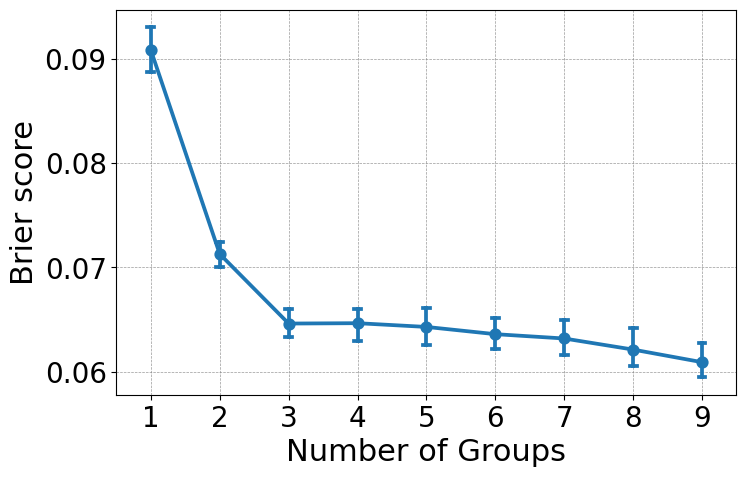}
        \caption{Brier loss (lower is better).}
        \label{fig:brier_x_numgroups}
    \end{subfigure}
    \caption{Mean area under the ROC curve (AUC) and Brier loss, which measures model calibration, when the EBM is trained on the top 1 through 9 groups of features ranked by group importance. Using just the top 3 groups yields nearly optimal performance, indicating feature selection may be performed effectively on a group level.}
    \label{fig:performance_x_numgroups}
\end{figure}

\paragraph{Accuracy for top groups:}
Based on the ranking of groups as defined in Table~\ref{tab:comparing_group_rankings} by our method, we compute the average test AUC and Brier score when trained on the top $1, \ldots, 9$ groups respectively (see Figure~\ref{fig:performance_x_numgroups}). Training with just the top three groups achieves a test AUC and Brier score almost comparable to using all nine groups. Figure~\ref{fig:performance_x_numgroups} shows that additional groups of features did not appear to significantly increase predictive power once (at least) the first three groups of features are used for training. However, Figure~\ref{fig:teaser} shows the remaining six groups are assigned a non-trivial group importance each. This implies they contain predictive power on their own, although they are, as a whole, sufficiently correlated with the groups of Life \& Trauma Events, Personality Traits, and Neuropsychological battery (and interactions between the features therein) to not improve the AUC.


\subsection{Healthcare Experiments}
In our healthcare case study, we evaluate the importance of community-level social determinants of health variables in predicting mortality after total hip arthroplasty. It is debated in the literature how important patient demographics, such as race, sex, and age are compared to broader, community-level social determinants of health (SDOH),  like healthcare access and socioeconomic status \cite{jha2005racial, lawrence1998estimates, mehta2019race}. In this section we compare an aggregate measure of SDOH, i.e. `community', to various individual factors related to demographics as well as the Elixhauser Comorbidity Index.

\subsubsection{Data}\label{sec:real-world-healthcare-data}
Our healthcare study utilizes data extracted from the Pennsylvania Health Care Cost Containment Council (PHC4) Database, covering the period from 2012 to 2018. This comprehensive dataset encompasses de-identified patient information, diagnostic and procedural codes, and financial data from 170 non-governmental acute care hospitals in Pennsylvania. Specifically, we focused on 140,092 patients who underwent elective primary Total Hip Arthroplasty (THA) according to validated International Classification of Diseases codes. After applying exclusion criteria such as inflammatory arthritis, non-elective admissions, and demographic omissions, our final cohort consisted of 105,336 patients. Key patient- and facility-level variables were extracted, including demographics and medical comorbidities indexed by the Quan adaptation of the Elixhauser Comorbidity Index. Unfortunately, certain granular, patient-level information such as a patient's income was unavailable, although it has been shown that socioeconomic status is very correlated with income and poses a reasonable proxy \cite{duncan2012socioeconomic}.

Community-level variables were an integral part of our analysis. These were extracted by geocoding the residential 5-digit zip codes of each patient to their corresponding census tract variables from the American Community Survey (ACS). We focused on variables known to influence THA outcomes, including social support indicators (percent living alone), immigration and acculturation factors (percent foreign-born, percent non-English speakers), digital literacy (percent with computer and internet access), socioeconomic status (median household income, percent uninsured), and education levels. Additionally, we included the National Walkability Index, given its relevance to osteoarthritis and post-arthroplasty outcomes. These variables collectively formed the `community factors' group in our analysis, allowing us to assess their aggregated impact alongside individual patient factors such as race.

\paragraph{Target variable} Our target variable is \textit{90-day mortality}, a critical measure in hospital settings due to its direct relevance to patient outcomes and healthcare quality. Mortality occurred in 0.3\% of cases.

\subsubsection{Results}
\begin{figure}[H]
    \centering
    \includegraphics[width=0.75\linewidth]{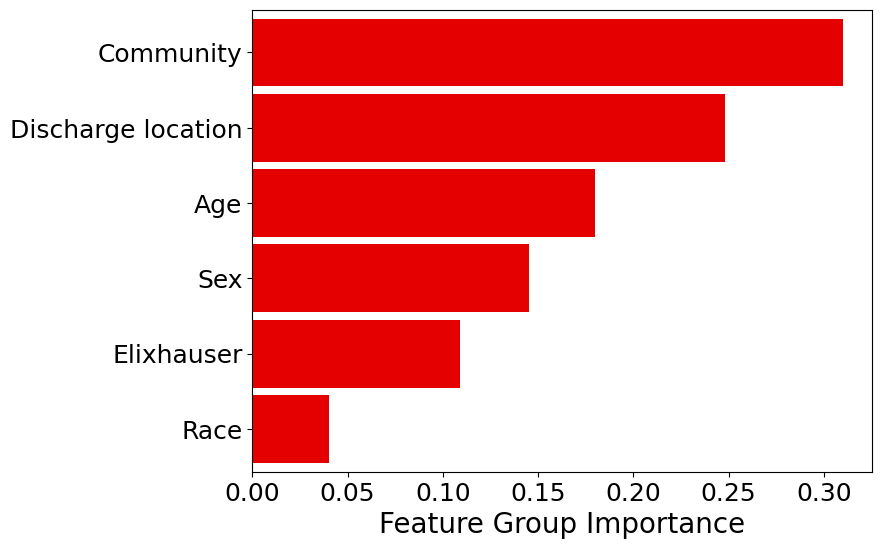}
    \caption{Group and individual feature importances for predicting 90-day mortality after total hip arthroplasty on the Pennsylvania Health Care Cost Containment Council (PHC4) database. Community-level social determinants of health collectively are more important than individual predictors such as age, sex, and comorbidities, traditionally considered the main risk factors \cite{jha2005racial, okike2019association, singh2014racial, belmont2014morbidity, dowsey2018impact, inacio2013sex}.}
    \label{fig:healthcare_gimp}
\end{figure}

Our EBM obtains an average test AUC of 0.76.
Similar to the neuroscience case study, we find that it is in fact a group of individually weak features that is driving the prediction of 90-day mortality, not necessarily individual features.  We compute the feature group importance of the group of community-based features, and contrast this in Figure~\ref{fig:healthcare_gimp} with the individual feature importance of race, sex, age, discharge location, and the Elixhauser Comorbidity Index.

Remarkably, the most important feature for predicting mortality in hip arthroplasty is the group of community-based measurements, carrying $\sim{}8$x the importance of Race, a well-recognized risk factor \cite{jha2005racial, okike2019association, singh2014racial}. Moreover, the community-level SDOH features are assigned more importance than sex and age combined, while these are also considered important risk factors for adverse outcomes after undergoing total hip arthroplasty \cite{belmont2014morbidity, dowsey2018impact, inacio2013sex}. The second most important feature is discharge location, which is an important predictor that could be informative of the amount of care available to a patient. It has been shown in the literature that patients discharged to a nursing or rehabilitation facility are less likely to have severe adverse events (e.g., mortality) compared to those who attempt to self-manage at home \cite{shah2017nonelective}.

These findings underscore the importance of considering social determinants of health in hip arthroplasty outcomes, and beyond in the healthcare system. Despite the tendency for social determinants to be highly correlated \cite{zimmerman2019trends, braveman2011social}, which often leads to low individual feature importance, their collective influence as a group is substantial. This reinforces the value of group importance analysis in revealing critical insights that might be obscured when focusing solely on individual feature importance.

\section{Conclusions}\label{sec:conclusion}
We propose a novel method to measure the importance of a group of features within a Generalized Additive Model (GAMs) by computing their mean absolute joint contribution. This approach naturally extends the concept of individual feature importance in GAMs, and remains meaningful when there are a lot of features. Our definition is computationally low-cost, requires no model retraining, and also offers a parallel with explained variation in statistics. We demonstrate its properties by applying a GAM-based Explainable Boosting Machine (EBM) to 2 real-world datasets (in neuroscience and healthcare) and 3 synthetic datasets. Our experiments reveal that group importance is particularly important for understanding the total signal in a group of correlated features, which can be overlooked when individual features are attributed low importance due to correlation. As such, group importance offers additional interpretability beyond individual feature importance in a wide range of applications.

\section*{Data Availability}
The data used for the real-world neuroscience experiment is made publicly available by the National Consortium on Alcohol and Neurodevelopment in Adolescence (NCANDA) \cite{ncanda2021} (data release \verb|NCANDA_PUBLIC_6Y_REDCAP_V04| \cite{pohl2022ncanda_public_6y_redcap_v04} and, for fMRIs,\\ \verb|NCANDA_PUBLIC_6Y_RESTINGSTATE_V01| \cite{pohl2022ncanda_public_6y_restingstate_v01}). The data for the real-world healthcare experiment is made publicly available by Pennsylvania Health Care Cost Containment Council (PHC4), and as of June 24, 2025, can be accessed at \url{https://www.phc4.org/request-custom-data/available-data/}.

 All data generated and analysed for the synthetic experiments are included in this published article and its Supplementary Information file.

\bibliography{sample}

\section*{Acknowledgements}
This work was generously supported by funding from the National Institute of Health (R01-AA005965, U01-AA017347, U24-AA021697) and Microsoft’s AI for Good Lab.

\section*{Author contributions statement}
T.B. led the analysis, conducted neuroscience and synthetic experiments, and wrote initial draft of manuscript.
L.S. and J.W. led the code development and integration into the InterpretML package.
Y.W. and B.M. provided clinical healthcare expertise, and conducted the experiments on social determinants of health.
J.D. prepared and analyzed the neuroscience data.
O.K. and F.B. provided clinical neuroscience expertise.
Q.Z. developed the fMRI segmentation methodology and led its application to the neuroscience experiments.
R.C. conceived of the presented idea and developed the theory.
K.M. collected and analyzed the neuroscience data.
Both R.C. and K.M. provided guidance, were closely involved in writing the manuscript, and led the direction of the project.
All authors contributed to discussion and reviewing the manuscript.

\section*{Additional information}
\textbf{Competing interests} The authors declare no competing interests.

\subsection*{Figures}

\clearpage
\subsection*{Tables}

\begin{table*}[ht]
    \centering
    \caption{Ranking of the group importances of the groups of features in the neuroscience case study using NCANDA data\cite{ncanda2021, pohl2022ncanda_public_6y_redcap_v04}. The 23 resting state fMRI (rs-fMRI) features and 131 demographic and behavioral measurements were grouped into nine disjoint groups \cite{paschali2022bridging, brown2015national}. The permutations-based ranking has been perviously reported \cite{paschali2022bridging} on NCANDA data excluding rs-fMRI brain network scores. The group permutation importance (GPI) metric \cite{plagwitz2022supporting} was also used for comparison. While there is a lot of similarity in the three rankings, our ranking seems most in line with the literature \cite{engle2005cognitive, hu2022depression, burt1995depression, brown1994cognitive, clark1985intellectual}. All three methods agree on the two most important groups, i.e., Life \& Trauma Events and Personality Traits, while Demographics and Alcohol \& Drug Use are the least important across all metrics.}
    \begin{tabular}{c c c c}
        \toprule
             Rank & Ours & Permutations \cite{paschali2022bridging} & GPI \cite{plagwitz2022supporting} \\ \hline \hline
             1 & Life \& Trauma & Personality & Personality  \\
             2 & Personality & Life \& Trauma & Life \& Trauma \\
             3 & Neuropsychology & Executive Function & Executive Function \\
             4 & Executive Function  & Sleep Patterns & Social Support \\
             5 & Social Support & Alcohol \& Drug Use & Neuropsychology \\
             6 & Sleep Patterns & Social Support & Sleep Patterns \\
             7 & Brain Networks & Neuropsychology & Brain Network \\
             8 & Demographics & Demographics & Demographics \\
             9 & Alcohol \& Drug Use &  & Alcohol \& Drug Use \\
        \bottomrule
    \end{tabular}

    \label{tab:comparing_group_rankings}
\end{table*}

\end{document}


\makeatletter
\vskip-36pt
{\raggedright\sffamily\bfseries\fontsize{20}{25}\selectfont \@title\par}
\makeatother

\appendix
\renewcommand\thesection{S\arabic{section}}
\renewcommand{\figurename}{Supplementary Fig.}
\renewcommand{\thefigure}{S\arabic{figure}}

\section{Group importances for synthetic data}\label{sec:app:group_imp_synth_data_A}
For notational convenience, we denote the uniform distribution over the set $\{a,b\}$ as $\mathcal{U}\{a,b\}$, although this typically denotes the set $\{a, a+1, \ldots, b\}$. The setup is the same as presented on synthetic experiments in the main manuscript.

\subsection{Additive signal with features $x$ and $-x$}\label{sec:app:exp1_x_and_-x}
We show that setting $z=-x$ in Section 3.1 leads to equivalent results when predicting label $y_i = 1$ if $x_i + \epsilon_i > 5$ and $0$ else, using an Explainable Boosting Machine (EBM). Mathematically, it is trivial that $t_{-x} = -t_x$ (e.g., if $t_x = 4$, then $t_{-x} = -4$), and so $f_{-x}(t_{-x}) = f_x(t_x)$ by point symmetry and the fact shape functions are zero-centered. This implies that
\begin{align}
    I_{\{x, -x\}} &= \frac{1}{|\mathcal{T}|} \sum_{t\in \mathcal{T}} |f_{x}(t_x) + f_{-x}(t_{-x})| \\
    &= \frac{1}{|\mathcal{T}|} \sum_{t\in \mathcal{T}} |f_{x}(t_x) + f_{x}(t_{x})| \\
    &= I_{\{x,x\}} =  2I_x,
\end{align}
averaging over samples $t=(t_x, t_{-x}) \in \mathcal{T}$.

Observe that $f_x(t_x)$ and $f_{-x}(t_{-x})$ contain equivalent information as visualized in Supplementary Fig.~\ref{fig:synthetic_data_exp_additive_signal+NEGATED}. Additionally, the features $x$ and $-x$ have the same feature importance, while their group importance is the sum of the individual importances as a result of their perfectly complementary information and shape functions,

\subsection{Correlated, conflicting signals with $\epsilon_i \overset{\text{iid}}{\sim} \mathcal{U}\{-2, 2\}$}\label{sec:app:group_imp_synth_data_eps2}
In the correlated, conflicting signals experiments in the main manuscript, all $\epsilon_i$ were sampled iid from a continuous uniform distribution $\mathcal{U}[a,b]$, and we have used scaling rather than clipping, which is also not uncommon in data science. We now explore what behavior the shape functions and group importances exhibit if we adopt the setting of correlated, conflicting signals setup from Section 3, but sample $\epsilon_i \overset{\text{iid}}{\sim} \mathcal{U}\{-2, 2\}$ and clip values to $[0,10]$. To construct conflicting signal, the target is defined as $y_i = 1$ if $x_i > z_i$ and $0$ else. The results are visualized in Supplementary Fig.~\ref{fig:synthetic_data_exp_noniid_eps2}.

Interestingly, the shape functions seem to represent (smoothed) step functions with a step size of 2, starting at 0. To see why such behavior might occur, consider $x_i \in [0,2]$. If $z_i=x_i+2 \in [2,4]$, then $y_i=0$ is guaranteed as $x_i < z_i$. If $z_i = 0$ (as $z_i=x_i-2 \in [-2,0]$ gets clipped to 0), then $y_i=1$, explaining the spike observed at $f_z(0)$. Thus, for all $x_i \in [0,2]$ where $z \neq 0$, it follows that $y_i=0$, resulting in the jump at $x=2$. This behavior seems to generalize for non-trivial values of $\epsilon_i$. Observe that this is \emph{not} purely an interaction term $f_{x,z}(\cdot, \cdot)$. In fact, if we allow the EBM to model $f_x(\cdot), f_z(\cdot), f_{x,z}(\cdot, \cdot)$ and enforce it to represent a functional ANOVA decomposition, we indeed recover similar shape functions $f_x(\cdot), f_z(\cdot)$.

Moreover, we observe a group importance less than the individual feature importances, which was expected given the correlation between $x$ and $z$ and their shape functions containing opposite signal.

\subsection{Correlated, conflicting signals with $\epsilon_i \overset{\text{iid}}{\sim} \mathcal{U}\{-0.1, 0.1\}$}\label{sec:app:group_imp_synth_data_eps0.1}
We adopt the setting from Supplementary Information S1.2 again, but sample $\epsilon_i \overset{\text{iid}}{\sim} \mathcal{U}\{-0.1, 0.1\}$ and clip values to $[0,10]$. Again, the target is defined as $y_i = 1$ if $x_i > z_i$ and $0$ else. The results are summarized in Supplementary Fig.~\ref{fig:synthetic_data_exp_noniid_eps0.1}.

\paragraph{Shape functions}
Perhaps surprisingly, the shape functions appear to have a near-flat center. We hypothesize this could be because the EBM discretizes the domain of $x,z$ into 256 bins by default and it's not able to learn fine effects. (With more bins, this problem indeed is resolved.) To illustrate the potential root cause of the near-flat center of the shape functions in Supplementary Fig.~\ref{fig:synthetic_data_exp_noniid_eps0.1}, consider samples $(x,z,y) = (5, 5.1, 0)$ and $(x,z,y) = (5.2, 5.1, 1)$, in which $z$ takes on a value of $5.1$ but has conflicting outcomes (0 and 1), possibly leading to a net-0 total contribution, suggesting that $\epsilon = \pm 0.1$ is small enough that it hinders the EBM from distinguishing between $z$ and $z + \epsilon$ on a relatively coarse partition of the domain.

As having small $\epsilon_i$ and a certain domain size is simply a matter of scale, we hypothesize that taking a finer partition of the domain space $[0,10]$ should lead to similar step functions as shown in Supplementary Information~\ref{sec:app:group_imp_synth_data_eps2}, namely with a step size of 0.1. Taking a finer partition at 4096 bins (instead of 256), we indeed recover more of a step-function-like shape function. These results highlight the importance of considering various (default) parameters while analyzing group importances and shape functions.

\paragraph{Group importance}
Observe that sampling $\epsilon_i \overset{\text{iid}}{\sim} \mathcal{U}\{-0.1, 0.1\}$ leads to strong correlation between features $x$ and $z$, and thus all pairs $(t_x, t_z) \in \mathcal{T}$ satisfy $|t_x-t_z| \leq 0.1$. Now, considering that the shape functions are additive inverses, we find that for all samples, $f_x(t_x) \approx -f_z(t_z)$ (again, because $x$ and $z$ carry contrary signal and shape functions are zero-centered), leading to a near-0 group importance $I_{x,z}$. Evidently, $I_{x,z} \to 0$ as $\epsilon_i \to 0$; this is shown in Figure 1(e).

\section*{Supplementary Figures}

\begin{figure}[h!]
    \centering
    \begin{subfigure}[t]{0.48\linewidth}
        \centering
        \includegraphics[width=\linewidth]{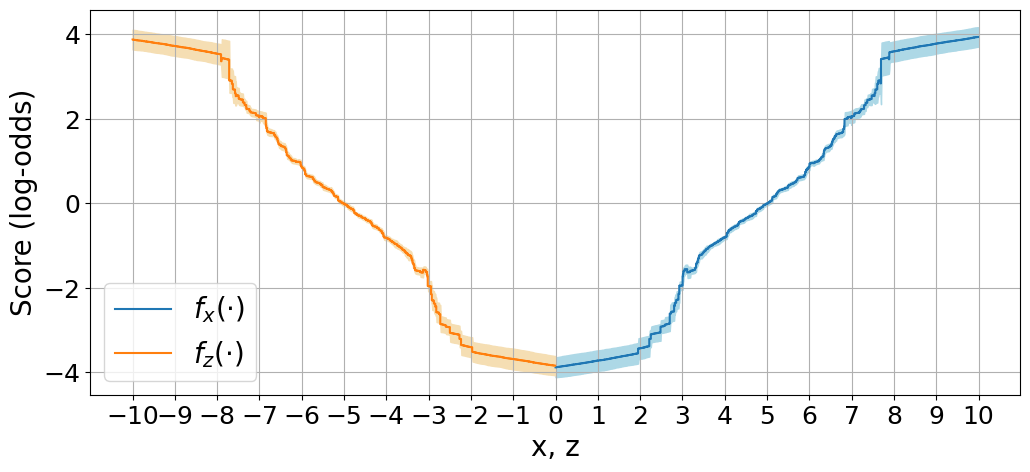}
        \caption{Shape functions $f_x(\cdot)$, $f_{-x}(\cdot)$ for features $x$ and $z=-x$. Note that $f_x(t_x) \approx f_{-x}(t_{-x})$.}
        \label{fig:synthetic_data_exp_additive_shapefuncs_NEGATED}
    \end{subfigure}%
    ~ 
    \begin{subfigure}[t]{0.48\linewidth}
        \centering
        \includegraphics[width=\linewidth]{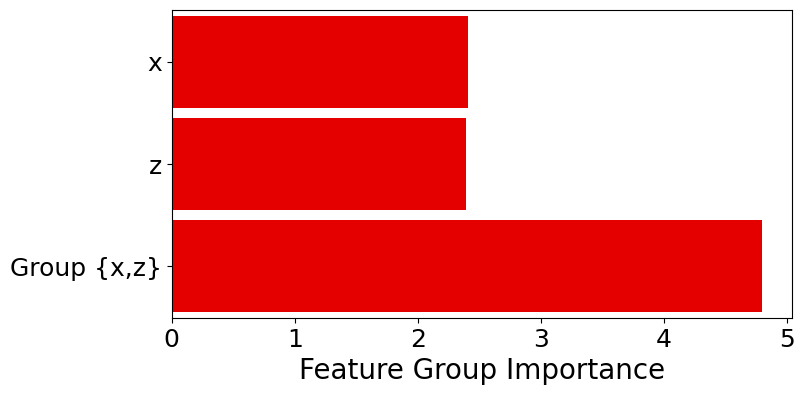}
        \caption{The feature importances display additive behavior.}
        \label{fig:synthetic_data_exp_additive_ft_imp_NEGATED}
    \end{subfigure}
    \caption{Synthetic-data experiment demonstrating additive behavior of correlated features x and z=-x, i.e., the feature z is a negated copy of x. This symmetry results in nearly identical shape functions. Notably, the group importance approximately equals the sum of individual importances features $x, -x$, showing the additive nature of group importances for features with additive signal. Individual feature importances are measured as the mean absolute contribution over all samples.}
    \label{fig:synthetic_data_exp_additive_signal+NEGATED}
\end{figure}

\begin{figure}[h!]
    \centering
    \begin{subfigure}[t]{0.5\linewidth}
        \centering
        \includegraphics[width=\linewidth]{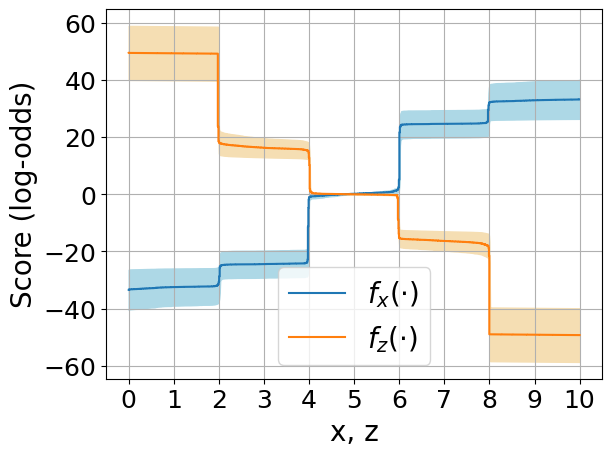}
        \caption{Shape functions $f_x(\cdot)$, $f_z(\cdot)$ for features $x$ and $z$, $\; z=x_i+\epsilon_i,\;  \epsilon_i \overset{\text{iid}}{\sim} \mathcal{U}\{-2, 2\}$.}
        \label{fig:synthetic_data_exp_conflicting_shapefuncs_noniid_eps2}
    \end{subfigure}%
    ~ 
    \begin{subfigure}[t]{0.5\linewidth}
        \centering
        \includegraphics[width=\linewidth]{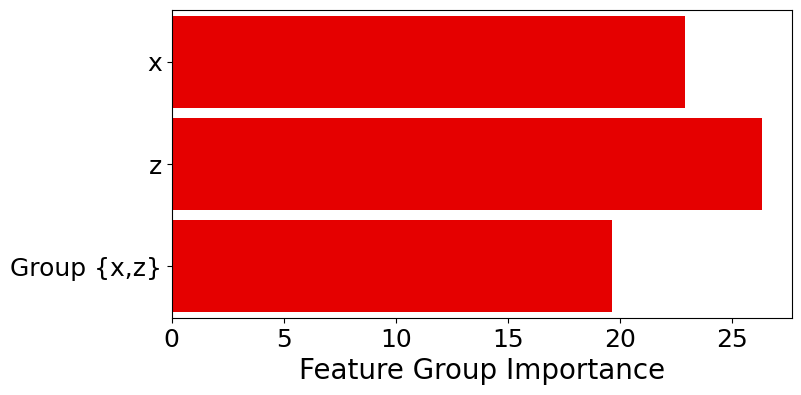}
        \caption{The feature importances are not additive for conflicting features.}
        \label{fig:synthetic_data_exp_conflicting_ft_imp_noniid_eps2}
    \end{subfigure}
    \caption{Synthetic-data experiment demonstrating conflicting signal from two correlated features x and z. The shape functions reflect similar stepwise contributions (as noise $\epsilon_i$ is fixed to $\pm 2$), and both features contribute opposingly to prediction.}
    \label{fig:synthetic_data_exp_noniid_eps2}
\end{figure}

\begin{figure}[h!]
    \centering
    \begin{subfigure}[t]{0.5\linewidth}
        \centering
        \includegraphics[width=\linewidth]{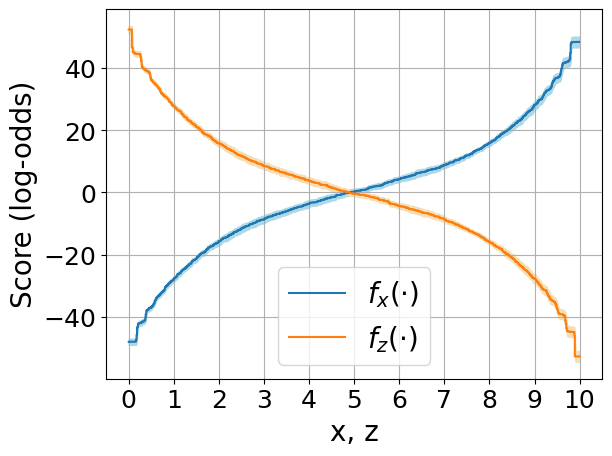}
        \caption{Shape functions $f_x(\cdot)$, $f_z(\cdot)$ for features $x$ and $z$, $\; z=x_i+\epsilon_i,\;  \epsilon_i \overset{\text{iid}}{\sim} \mathcal{U}\{-0.1, 0.1\}$.}
        \label{fig:synthetic_data_exp_conflicting_shapefuncs_noniid_eps0.1}
    \end{subfigure}%
    ~ 
    \begin{subfigure}[t]{0.5\linewidth}
        \centering
        \includegraphics[width=\linewidth]{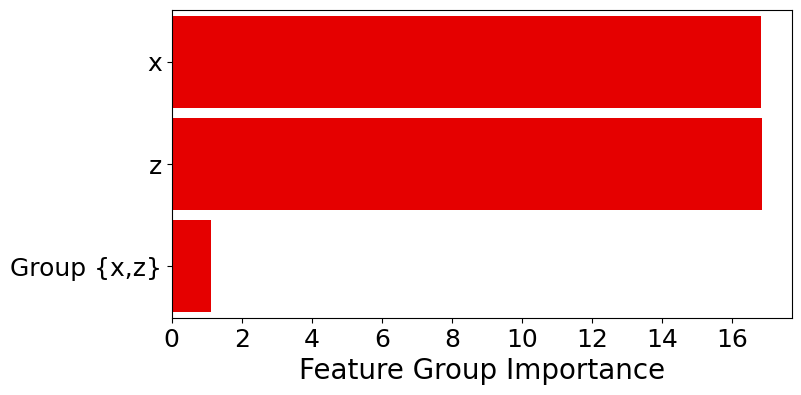}
        \caption{The individual feature importances are equal as a result of symmetry. Because feature values of $x$ and $z$ are always close together (as $|\epsilon_i| = 0.1$ for all $i$), and contain opposing signal, their group importance is almost zero.}
        \label{fig:synthetic_data_exp_conflicting_ft_imp_noniid_eps0.1}
    \end{subfigure}
    \caption{Synthetic-data experiment demonstrating conflicting signal between two correlated features x and z. The shape functions are near inverses, leading to equal individual importances. On the other hand, the group importance goes to zero as the feature values of x and z get closer together and their shape functions become near inverses.}
    \label{fig:synthetic_data_exp_noniid_eps0.1}
\end{figure}